\documentclass[a4paper,twoside,11pt]{article}

\usepackage[USenglish]{babel} 
\usepackage[T1]{fontenc}

\usepackage{graphicx}
\usepackage[labelsep=endash, figurename=Figure, tablename=Table, textfont=it]{caption}
\usepackage{subcaption}
\usepackage{float}  
\usepackage{hyperref, wasysym}
\usepackage{multicol}
\usepackage{wrapfig}

\usepackage{fancyhdr}
\usepackage[top=2.5cm, bottom=2.5cm, left=2.5cm, right=2.5cm]{geometry} 
\usepackage[usenames,dvipsnames]{pstricks}
\usepackage[round,comma,authoryear]{natbib}
\usepackage{titling}
\usepackage{enumitem}

\definecolor{darkblue}{rgb}{0,0,0.5}
\hypersetup{colorlinks=true,linkcolor=black,citecolor=darkblue,urlcolor=darkblue}

\newcommand{\eM}[0]{eMARLIN}
\newcommand{\eMP}[0]{eMARLIN+}
\newcommand{\eML}[0]{eMARLIN-LSTM}
\newcommand{\eMT}[0]{eMARLIN-Transformer}

\makeatletter
\let\ps@plain\ps@fancy
\makeatother

\begin{document}


\title{\bf Mitigating Partial Observability in Adaptive Traffic Signal Control with Transformers}
\author{X. Wang$^{a, *}$, A. Taitler$^{b}$, S. Sanner$^{a}$ and B. Abdulhai$^{a}$}
\date{\today}

\pretitle{\centering\Large}
\posttitle{\par\vspace{1ex}}

\preauthor{\centering}
\postauthor{\par\vspace{1ex}
$^{a}$ University of Toronto, Toronto, Canada\\
cnxiaoyu.wang@mail.utoronto.ca, baher.abdulhai@utoronto.ca, ssanner@mie.utoronto.ca\\
$^{b}$ Ben-Gurion University of the Negev, Be'er Sheva, Israel\\
ataitler@gmail.com\\
$^{*}$ Corresponding author

\vspace{1ex}\it
Extended abstract submitted for presentation at the Conference in Emerging Technologies in Transportation Systems (TRC-30)\\
September 02-03, 2024, Crete, Greece\\
\vspace{1ex}

}

\maketitle
\vspace{-1cm}
\noindent\rule{\textwidth}{0.5pt}\vspace{0cm}

\fancypagestyle{firststyle}{
\lhead[]{}
\rhead[]{}
\lfoot[TRC-30]{TRC-30}
\rfoot[Original abstract submittal]{Original abstract submittal}
\cfoot[]{}
}
\thispagestyle{firststyle}

\pagestyle{fancy}
\fancyhead{}
\fancyfoot{}
\renewcommand{\headrulewidth}{0pt}
\renewcommand{\footrulewidth}{0pt}
\setlength{\headheight}{15pt}
\lhead{X. Wang, A. Taitler, S. Sanner and B. Abdulhai}
\rhead[\thepage]{\thepage}
\lfoot[TRC-30]{TRC-30}
\rfoot[Original abstract submittal]{Original abstract submittal}
\cfoot[]{}


\section{INTRODUCTION}
Traffic signal control in urban areas plays a pivotal role in managing transportation, reducing congestion, and enhancing safety and sustainability \citep{grote2016including}. While traditional fixed-timing systems struggle to adapt to changing traffic conditions, Adaptive Traffic Signal Control (ATSC) systems use dynamic algorithms to adjust signal timings based on real-time data \citep{el2013multiagent, wang2023critical}, aiming to optimize traffic flow and minimize delays.

Reinforcement Learning (RL) has emerged as a promising approach to enhancing ATSC systems. RL algorithms enable controllers to learn optimal policies by interacting with the environment. One notable RL-based ATSC framework is eMARLIN \citep{wang2023emarlin}. eMARLIN introduces a lightweight distributive collaboration architecture, allowing a learned communication signal to be transmitted between neighbouring intersections while conserving the communication bandwidth.



However, a significant challenge in ATSC arises from partial observability (PO) in traffic networks. Agents rely on local sensors that can fail to accurately report system states or restricted by their maximum detection range. PO leads to uncertainty and complexity, hindering the control effectiveness. To address this, eMARLIN+ \citep{wang2023emarlinplus} extends eMARLIN by incorporating historical information using a Long Short-Term Memory (LSTM) network. It communicates abstracted local observation histories among agents, aiming to mitigate the effects of partial observability and improve overall system performance. However, LSTM's training throughput poses challenges.


Transformers, known for their success in sequence modelling tasks, including natural language processing (NLP), offer a promising alternative \citep{vaswani2017attention}. This paper explores integrating Transformer-based controllers into eMARLIN systems to effectively address partial observability. We propose strategies to enhance training efficiency and effectiveness, presenting results on real-world scenarios and comparing with strong baselines. The Transformer-based model demonstrates improved coordination capability by capturing significant events from historical data for better control policies.


\section{TRANSFORMERS PRELIMINARIES}



\begin{wrapfigure}{R}{0.28\textwidth}
    \centering
    \includegraphics[width=0.25\textwidth]{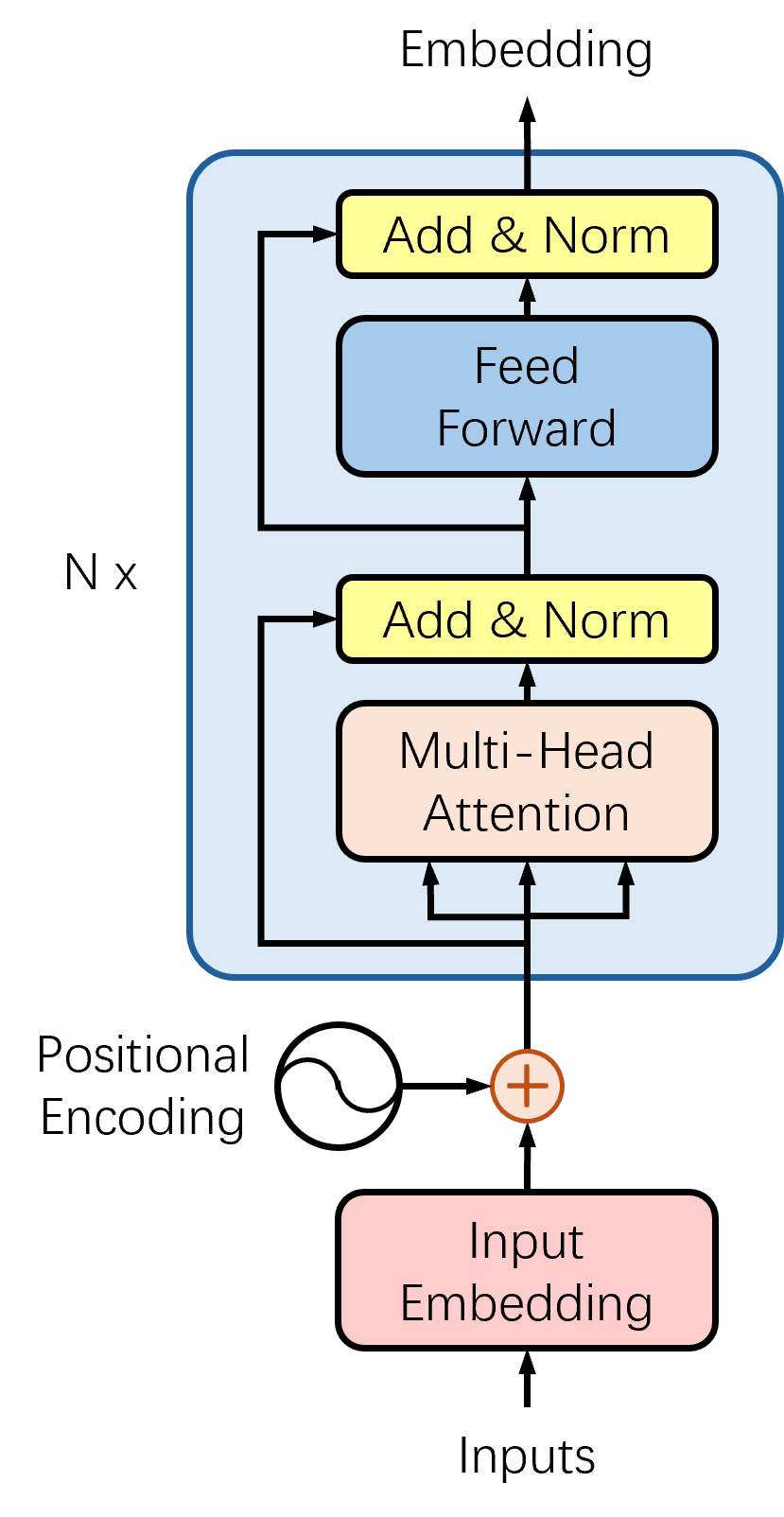}
    \caption{The architecture of a Transformer encoder model. }
    \vspace{-20pt}
    \label{fig:transformer-encoder}
\end{wrapfigure}


Attention mechanisms in machine learning assign weights to focus on specific parts of inputs, providing powerful feature abstraction capabilities. Transformers \citep{vaswani2017attention} are a significant advancement, demonstrating strong performance across various fields.

The Transformer encoder (Figure \ref{fig:transformer-encoder}), a key component, introduces two key innovations: positional encoding and the Transformer block. Positional encoding, typically using sinusoidal functions, provides order awareness to the embedded inputs, allowing Transformers to process sequences in parallel. This method is more scalable and interpretable than traditional recurrent architectures, offering a more efficient solution to long-term memory challenges.

The Transformer block's central innovation is its scaled dot-product attention mechanism, which calculates attention scores between linearly transformed input slices to produce a weighted combination of features. Multiple attention heads enhance understanding by focusing on different aspects of the input simultaneously. This multi-head approach improves scalability and adaptability, making Transformers highly effective for handling complex tasks in various domains.

\section{METHODOLOGY}
\subsection{The Underlying Model}
eMARLIN+ \citep{wang2023emarlinplus} is a distributed ATSC coordination algorithm based on the Deep Q-Network (DQNs) \citep{mnih2013playing}. It formulates the generic ATSC problem as a Partially Observable MDP (POMDP), where state variables define network dynamics, and observations capture sensor data.
To reduce the model complexity, eMARLIN+ factorizes the POMDP by introducing a Dominance Relaxation assumption, prioritizing immediate neighbourhood influences over distant ones under second-level operation. This relaxation factorizes the joint conditional distribution, that represents the global dynamics, and streamlines the model. This relaxation allows eMARLIN+ agents to learn only from neighbouring information, rather than the complex global data. Finally, \eMP~ decouples the global reward down to the sum over individual intersection delays, decoupling the joint problem into separate learnable and manageable sub-problems.

In this setting, each of the agents observes locally, receives a local reward, and maximizes this surrogate target directly. The model is designed to be:

\begin{itemize}[itemsep=4pt]
    \item \emph{States:} The state of the system $s \in \mathcal{S}$ contains all the information related to the network, including vehicle positions, velocities, traffic lights status, etc. This information in general is hidden and defines the underlying dynamics of the problem.
    \item \emph{Observations:} The local observation $o_i \in \mathcal{O}_i$ directly sensed by an agent consists of the count of vehicles and the count of queued vehicles (driving below a threshold speed, $2$ m/s used in this work) in each incoming lane within the detection range, as well as the current phase index and its elapsed duration.
    \item \emph{Actions:} The action $a_i \in \mathcal{A}_i$ represents the \textit{desired phase} in the next time-step. The traffic light EXTENDs the current phase if it is the same as the action, or CHANGEs to the selected phase by going through the corresponding yellow change and red clearance intervals. A policy constraint permits only those valid transitions under the standard NEMA two-ring-two-barrier scheme \citep{wang2023emarlin}. Minimum and maximum green time constraints are also applied.
    \item \emph{Rewards:} The local reward $r_i \in \mathcal{R}_i$ is the agent sensed total queue length at the intersection.
\end{itemize}

\subsection{eMARLIN-Transformer}
\begin{wrapfigure}{R}{0.4\textwidth}
    \centering
    \includegraphics[width=0.35\textwidth]{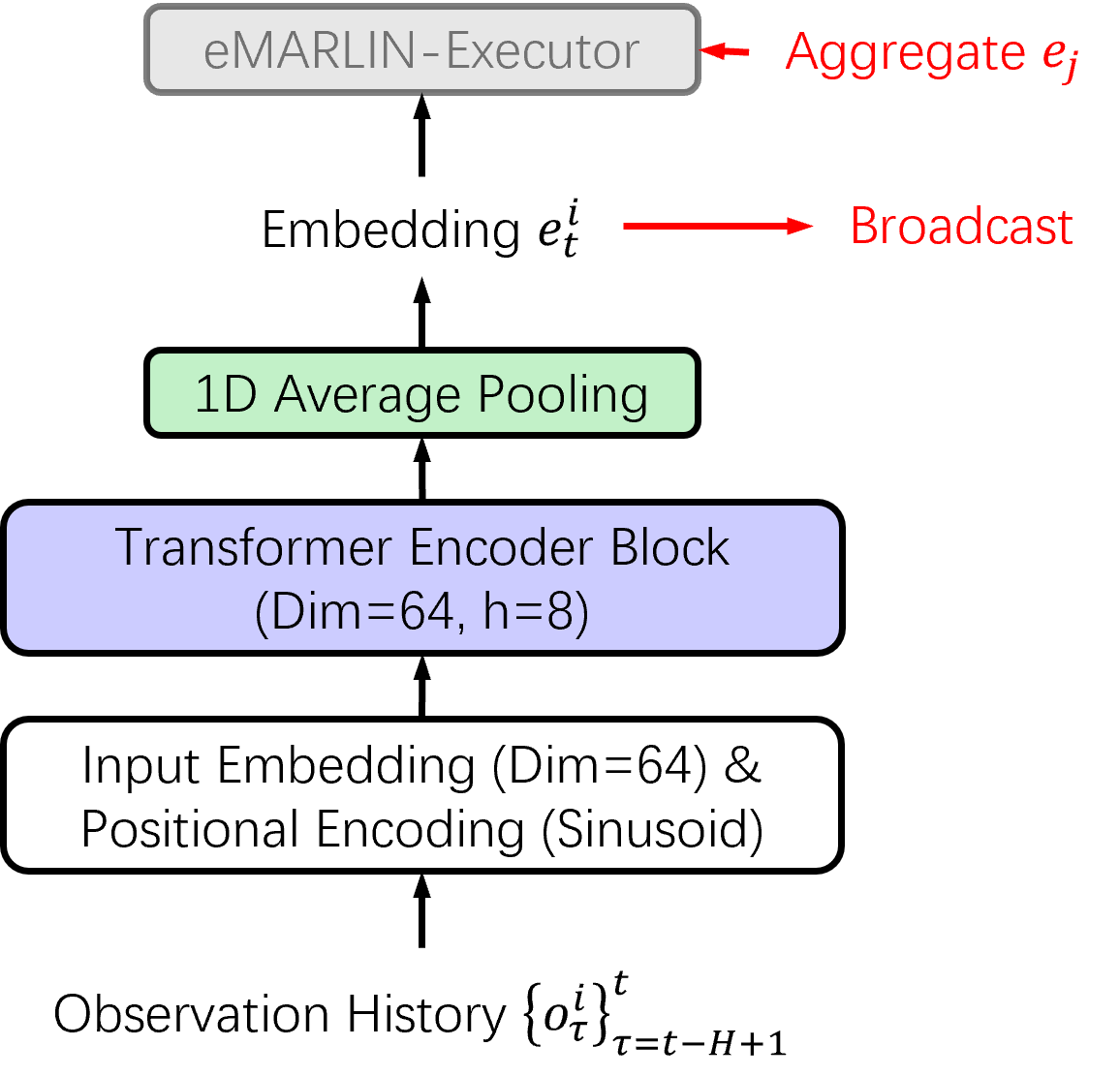}
    \caption{The architecture of Encoder in eMARLIN-Transformer.}
    \vspace{-20pt}
    \label{fig:eMT-encoder}
\end{wrapfigure}

To address partial observability caused by sensor limitations, we propose eMARLIN-Transformer, an integration of Transformers into the eMARLIN architecture. This extension replaces the fully connected network in the eMARLIN Encoder with a Transformer-based design and augments input with observation history. Unlike eMARLIN+, which uses LSTM and relies on recurrent hidden states, eMARLIN-Transformer leverages attention mechanisms for parallelized computations, improving efficiency and capturing temporal information more effectively. Note that for isolated intersections, eMARLIN-Transformer operates as DQN-Transformer.


\subsection{Implementation details}
RL and Transformers offer potential in traffic management tasks but face challenges like high memory usage and latency. To stabilize Transformer learning for ATSC, selecting a proper model structure is crucial (Figure \ref{fig:eMT-encoder}).

Handling historical information in Transformers requires careful data order and alignment. Our approach involves randomly selecting a recorded episode and a time slice from the replay buffer, retrieving the history up to a certain length. To maintain semantic alignment, we reverse sampled sequences before padding and positional encoding, emphasizing the current time slice.

Padding with zeros can confuse learning agents, as zeros might represent either padding or a zero-valued feature. To address this, we use padding masking, which ensures that attention mechanisms do not misinterpret padded sequences. This masking technique is also applied at the pooling layer after the attention block to ensure clarity throughout the learning process.

\section{RESULTS}
We modelled a neighbourhood surrounding the intersection of Yonge Street and Steeles Avenue in Toronto, Canada using the Aimsun simulator. City signal timing plans, obtained from Toronto, follow the standard NEMA scheme with semi-actuated control \citep{NEMA-actuated}. RL agents run on five intersections along Yonge Street, while the remaining follow city signal timing plans.

\begin{table*}[htbp]
\begin{center}
\small
\caption{eMARLIN-Transformer outperforms all baselines on the Toronto test-bed. Numbers reported in the table are the episodic delay (s) of each intersection.}
    \begin{tabular}{ c|cccccc }
        \hline
        & North2 & North1 & Yonge-Steeles & South1 & South2 & Sum \\
        \hline
        City plan   & 2,120 & 77,222 & 465,500 & 6,091 & 49,965 & 600,898 \\
        PressLight  & 23,590 & 101,091 & 276,824 & 60,180 & 115,070 & 576,755 \\
        DQN        & 7,169 & 61,029 & 197,544 & 40,999 & 50,288 & 357,029 \\
        eMARLIN & \textbf{1,027} & 60,800 & 174,464 & 8,174 & 48,170 & 292,635 \\
        DQN-LSTM &	1,134 & 65,033 & 155,048 & 4,499 & 38,406 & 264,120 \\
        \eML & 1,407 & 51,228 & 156,408 & \textbf{3,996} & 29,706 & 242,745 \\
        \eMT & 1,416 & \textbf{40,491} & \textbf{136,731} & \textbf{4,126} & \textbf{22,747} & \textbf{205,511} \\
        \hline
    \end{tabular}
    \label{tab:NY-eMT}
\end{center}
\end{table*}


The results in Table \ref{tab:NY-eMT} highlights the effectiveness of \eMT~ in coordination performance compared to other methods. Specifically, \eMT~ reduces delay time by $65.8\%$ compared to a semi-actuated city plan, indicating a substantial improvement in traffic flow.

When compared to the naive \eM, which does not attempt to address the PO issue, \eMT~ achieves a $30.8\%$ improvement in delay time, demonstrating the importance of accounting for sensing limitations and utilizing historical data. Additionally, \eMT~ outperforms \eML, the previous state-of-the-art approach that uses LSTM to handle sequential data, with $16.6\%$ greater delay time savings, underscoring the advantages of Transformer-based models in addressing PO issues.

These results underscore the Transformer's capacity to capture sequential information from historical observations, enhancing multi-agent collaboration in ATSC, leading to more efficient traffic management and reduced congestion.

\section{CONCLUSION AND DISCUSSION}
In this work, we presented Transformers as an alternative encoder for capturing sequential information in traffic scenarios, providing advantages over LSTM networks, particularly in handling PO. We integrated Transformers into the eMARLIN architecture to create eMARLIN-Transformer. The implementation details were discussed, highlighting the Transformer’s improved performance compared to other RL baselines. Experiments in the North York test-bed demonstrated that the Transformer-based encoder effectively addresses PO issues, enhances agent coordination, and leads to better control performance.


\begin{small}
\begin{sloppypar} 
\bibliographystyle{authordate1} 

\setlength{\bibsep}{0pt}

\bibliography{main}

\end{sloppypar}
\end{small}


\end{document}